\documentclass[a4paper,fleqn]{cas-dc}
\usepackage[numbers]{natbib}
\usepackage{caption}    
\usepackage{subfig}
\usepackage{algorithm}
\usepackage{algpseudocode}

\def\tsc#1{\csdef{#1}{\textsc{\lowercase{#1}}\xspace}}
\tsc{WGM}
\tsc{QE}
\tsc{EP}
\tsc{PMS}
\tsc{BEC}
\tsc{DE}
\begin{document}

\let\WriteBookmarks\relax
\def\floatpagepagefraction{1}
\def\textpagefraction{.001}
\shorttitle{$\sigma^2$R Loss}
\shortauthors{La Grassa R et~al.}


\title [mode = title]{$\sigma^2$R Loss: a Weighted Loss by Multiplicative Factors using Sigmoidal Functions}



\author[1]{Riccardo La~Grassa}[orcid=0000-0002-4355-0366]
\author[1]{Ignazio Gallo}[orcid=]
\author[1]{Nicola Landro}[orcid=]

\address[1]{University of Insubria, Department of Theoretical and Applied Science, Italy}


\begin{abstract}
In neural networks, the loss function represents the core of the learning process that leads the optimizer to an approximation of the optimal convergence error. 
Convolutional neural networks (CNN) use the loss function as a supervisory signal to train a deep model and contribute significantly to achieving the state of the art in some fields of artificial vision.
Cross-entropy and Center loss functions are commonly used to increase the discriminating power of learned functions and increase the generalization performance of the model. Center loss minimizes the class intra-class variance and at the same time penalizes the long distance between the deep features inside each class. 
However, the total error of the center loss will be heavily influenced by the majority of the instances and can lead to a freezing state in terms of intra-class variance.
To address this, we introduce a new loss function called sigma squared reduction loss ($\sigma^2$R loss), which is regulated by a sigmoid function to inflate/deflate the error per instance and then continue to reduce the intra-class variance.
Our loss has clear intuition and geometric interpretation, furthermore, we demonstrate by experiments the effectiveness of our proposal on several benchmark datasets showing the intra-class variance reduction and overcoming the results obtained with center loss and soft nearest neighbour functions.
\end{abstract}




\maketitle

\section{Introduction}
Nowadays deep learning is taking on an ever clearer form in terms of different elements which is composed and in their possible configurations, achieving very good results in different tasks as Computer Vision~\cite{krizhevsky2012imagenet}, Speech Recognition, multimodal methodologies, Natural Language Processing (NPL) and hybrid models like OCMST~\cite{la2020ocmst}~\cite{grassa2020dynamic}, autoencoder OCSVM~\cite{andrews2016detecting}, autoencoder based on KNN~\cite{song2017hybrid}, autoencoder SVDD~\cite{kim2015deep}, One-Class Neural Networks  (OCNN)~\cite{chalapathy2018anomaly,ruff2018deep,perera2019learning}, and many other fields.
Convolutional Neural Networks (CNN) has been widely used in image recognition, such as face recognition and image classification achieving the state-of-the-arts in most cases. 
The general capability to generalize discriminative deep features is due to different elements inside the convolutional neural networks. 
A fundamental element of the learning process is called the loss function, which will be minimized by techniques such as the descent of the stochastic gradient to favour the synchronization of neurons and therefore be able to solve a particular problem.
However, the general characteristics of the loss function and other CNN elements are still poorly understood and some key concepts are still seen as a black box~\cite{choromanska2015loss}.
In literature, there are various types of loss functions applied mainly in the last layer of the neural model. 
The most popular are for example the Softmax cross-entropy loss function and similar~\cite{asoftmax,cosface}, the Hinge loss function~\cite{cortes1995support}, the Ramp loss function~\cite{ramp}, the Additive Angular Margin Loss~\cite{arcface}, etc.. 
Other loss functions are applied into features layers before the last layer or in some cases in other parts of deep networks. 
The effect of these loss functions is reflected in the hyper-dimensional space, for example by improving the discriminatory ability, or by trying to increase the distance between the classes and reducing the variance within each class, using the training samples~\cite{qi2017contrastive}.
The latter is the objective of a known loss function called center loss~\cite{wen2016discriminative} which from experimental results demonstrate its effectiveness and usability to achieve, together with cross-entropy loss, the state of the art in various activities such as facial recognition and verification problems.

Recent papers such as~\cite{frosst2019analyzing}, introduce a new loss function that can measure the entanglement on the labelled data by establishing how close the pairs of instances of the same class are.
By decreasing the entanglement around the data with the loss function called soft nearest neighbour, they can get more than one cluster and not have a forced convergence at a single centroid for each class as is the case for the center loss function.

In this paper, we introduce a new loss function called $\sigma^2$R Loss, whose main goal is to inflate/deflate the error generated by each instance of the training set.
This proposed loss function is regulated by sigmoidal functions automatically configured by hyper-parameters learned during the training process.
The main motivation that led us to propose this new loss function comes from the characteristics of the center loss function.
Observing the center loss function~\cite{wen2016discriminative} and variation of it, the ability to reduce the variance between deep features within a class decreases when the instances of features close to the class centroid are many compared to those that are very far from the same centroid.
The main reason is that the Euclidean distance for each training sample is not weighted and therefore if there are only a few samples far from the class center while all the others are close, the total error will be low because strongly influenced by the number of nearby instances.
To address this problem, in our proposal, we apply a weight as a function of the distance from the class center with the effect of increasing the error for the furthest points, to enhance the discriminating ability of CNN models.
In this paper, we choose to use a trainable sigmoid function as a multiplicative factor which in turn depends on a few hyper-parameters.

In summary, the main contributions of this paper are listed as follows:
\begin{itemize}
    \item First, we propose a new loss function based on the center loss function, useful to minimize the variance of the deep features within each class to improve the generalization ability of neural networks.
    \item Second, we conduct many experiments on well-known benchmark datasets demonstrating the effectiveness and high usability of our proposal on different CNN architectures.
    \item Third, all our source code is available on GitLab~\cite{s2R_link}, to allow the community to reproduce our results, from the training of the networks, until the statistical analysis.
\end{itemize}

In the next section, we introduce some of the loss functions directly related to what we propose in this paper, to facilitate the understanding of our proposal.
The following sections will introduce the details of the proposed approach and the experiments conducted to demonstrate its characteristics and strengths.

\section{Related Work}
Cross-entropy based softmax loss is a well-known loss function widely used in machine learning to discriminate classes in classifications problem or for feature learning.
It encourages the separability of features but people realized that the cross-entropy loss is not sufficiently effective to learn feature with a large margin for some problems like for example face recognition.
Constructing highly efficient loss function to increase the discriminative power is not a trivial problem.
Many solutions have been proposed in the literature to increase the generalization performance of models using specific loss functions.
Most of these loss functions are applied to the second-last layer of a deep network, or into the last layer as cross-entropy loss is usually applied.
Center loss~\cite{wen2016discriminative}, Contrastive Center loss~\cite{qi2017contrastive}, Triplet Center loss~\cite{he2018triplet} are some loss functions commonly used to improve the approximation of the optimal solution, compared to what cross-entropy can do.
These loss functions are Euclidean-distance-based loss~\cite{mei2018deep} and they focus to compress intra-class variance and enlarge inter-class variance. 
In~\cite{kulesza2015low}, the authors introduce a loss function orientated to resolve a problem in the domain of text prediction that uses a weighted function depending by the length of a string $x$ as $w(x)=\frac{1}{2^{|x|}}$ and they use the squared Euclidean distance between two probabilities of observed letters in a text document to minimize the error.
Although this is similar to our proposed approach because they use a weighted function, we differentiate our paper with it for the following reasons.
We use squared Euclidean distance applied to features layers from a Convolutional Neural Network. 
The weighted functions used are sigmoids (one per class) with the logistic growth rate variables inserted into the learning process.
Secondly, we inflate/deflate the error per instance considering standard deviations and distances combined. Finally, we applied our solution to computer vision tasks using well-known images dataset.

In the next subsections, we briefly describe the loss functions that belong to the Euclidean-distance category~\cite{mei2018deep} and we highlight their characteristics. 
All the loss functions that we report below also have among their objectives the reduction of the intra-class variance. 
In our paper, we will not compare with all these loss functions but only with those that share the same objective.

\subsection{Center Loss}
While the cross-entropy loss function focuses on classification errors by attempting to minimize them, the center loss minimizes the distance of each class point in the feature space from its center.
The effect of cross-entropy loss in the features space is to separate the features of different classes but not to maximize the margin between the classes.
The cross-entropy loss is not able to reduce the variance within the class, always analyzing the features space.
Center loss is widely used to solve this task decreasing the intra-class variance and increasing the general performance of the model. 
It is used jointly with cross-entropy loss function overcoming the state-of-the-art in most classification problems. 
For example, \cite{ghosh2018understanding} shows that for datasets with a large number of classes but a small number of samples per class, the combination of cross-entropy loss and center loss works better than either of the losses alone.

Formally, the center loss is computed as: \begin{equation}\label{equ:center_loss}
    \mathcal{L}_C=\frac{\lambda}{2}\sum\limits_{i=1}^m {|| x_i-c_{y_i}||}_2^2
\end{equation}
where $c_{y_i} \in \mathbb{R}^d$ denotes the center of the $y_i$-th class in the deep features space, $x_i$ is an instance of the class $y_i$ and $m$ is the batch size. 

\subsection{Soft nearest neighbor loss}
Recently, a novel loss function is introduced in~\cite{frosst2019analyzing}, called Soft Nearest Neighbor. It measures entanglement over labeled data and it is defined as follow:
\begin{equation}
\label{eq:snnloss}
\mathcal{L}_{sn} = -\frac{1}{m} \cdot \sum\limits_{i=1}^m log\left(
\frac
{\sum\limits_{j \in [1,m], j \neq i, y_i = y_j} e^{-\frac{||x_i - x_j||^2}{T}} }
{\sum\limits_{k \in [1,m], k \neq i} e^{-\frac{||x_i - x_k||^2}{T}}}
\right)
\end{equation}
where $T$ is the temperature (a non-learned parameter), $m$ is the batch size, $x_i$ is the output of layer in which this loss function is applied and $y_i$ is the label of the $i$-th instance.
Intuitively this loss brings features of the same class closer together while moving away those of other classes.
An effect of this loss is to create class-independent clusters and not necessarily to converge all instances in a single point.
Although the results reported by authors demonstrate improvements in terms of accuracies, they do not compare with center loss (whose goal is also to minimize the intra-class variance).

\subsection{Inter-class distance maximization loss functions}

This section groups all the loss functions which, in addition to minimizing the intra-class variance, also focus on maximizing the inter-class distance.
Since this last objective is not our goal, we describe only these loss functions, but we will not compare with them in our experiments.

The \textbf{Git loss} function~\cite{calefati2018git} maximizes the distance between deeply learned features belonging to different classes (push) while keeping features of the same class compact (pull) using the Eq.~\ref{eq:gitloss} that can be simplified with Eq. \ref{eq:semplifiedGitloss}
\begin{equation}
    \label{eq:gitloss}
    \mathcal{L}_G = \mathcal{L}_{xent} + \frac{\lambda_c \mathcal{L}_{c}}{2} + \lambda_g \cdot \sum_{i,j=1,i \neq j }^m \frac{1}{1 + || x_i - c_{y_j} ||_2^2}
\end{equation}

\begin{equation}
    \label{eq:semplifiedGitloss}
    \mathcal{L}_G = \mathcal{L}_{xent} + \sum_{i,j=1,i \neq j }^m  -\frac{2 \cdot (x_i - c_{y_j})}{(1 + (x_i - c_{y_j})^2)^2} 
\end{equation}
where $m$ denotes the number of training samples of a batch, $x_i\in \mathbb{R}^d$ is the $i$-instance of the training, $c_{y_i} \in \mathbb{R}^d$ denotes the center of the $y_j$-th class in the deep features space, $\mathcal{L}_{xent}$ represent the cross-entropy loss function.
The results reported in~\cite{calefati2018git} show a higher accuracy obtained thanks to the reductions in terms of intra-class and inter-class distance.

Ce Qi \textit{et al}~\cite{qi2017contrastive} with the proposed \textbf{contrastive center loss} extend the center loss to simultaneously reduce the intra-class variance and increase the inter-class distance jointly with cross entropy loss. More formally,
\begin{equation}
\label{equ:constrastive_center_loss}
    \mathcal{L}_{CT}=\frac{1}{2}\sum\limits_{i=1}^m {\frac{|| x_i-c_{y_i}||_2^2}{(\sum_{j=1,j \neq y_i}^k {|| x_i-c_{j}||}_2^2)+\delta}}
\end{equation}
where $m$ denotes the number of training samples in a batch, $x_i\in \mathbb{R}^d$ is the $i$-th instance of the training with $d$ dimension, $k$ is the number of classes and $\delta$ is a constant to avoid division by 0.
Results show the effectiveness of this loss function to better generalize the deep features and obtain more reduction in terms of intra-class variance and more separability in terms of inter-class distance.


The \textbf{triplet center loss}~\cite{he2018triplet}  is calculated through the use of triplets of instances and is intended to bring together the elements of the same class and at the same time increase the distance between the classes.
The triplet is composed as $(x_a^i, x_+^i, x_-^i)$ where $x_a^i$ is called anchor and have the same class of $x_+^i$ that is called a positive sample, instead $x_-^i$ is the nearest sample to the anchor $x_a^i$ of another class and it is called negative sample. 
Using this approach, for each positive instance, triplets for all negative samples should exist. But for efficiency reasons, only the triplets containing the closest negative instance are selected.
So triplet center loss can be computed as described in the following Eq.~\ref{eq:triplet}.
\begin{eqnarray}\label{eq:triplet}
L_{tpl} & = & \sum_{i=0}^N max(0, m + D(f(x_a^i), f(x_+^i))- \nonumber \\ 
        &  & - (D(f(x_a^i), f(x_{-}^j))))
\end{eqnarray}
where $f(x)$ represents the output of the feature layer of the neural network taking in input $x$; $D(a, b)$ is the selected distance measure between two instances $a$ and $b$; $N$ represents the number of triplet int the current batch, finally $D(x_a^i, x_b^j)$ with $i \neq j$ represents the nearest distance from a sample of another class from the anchor, 
and $m$ is a hyper-parameter that represents the margin between two instances belonging to two different classes.

The \textbf{contrastive loss}~\cite{lian2018speech} is similar to the triplet center loss. 
It brings all the points of the same class (paired) closer together and moves them away from those of other classes (not paired). 
So for all couple of instances $r_a$ and $r_b$  available in a batch, the loss function is described as reported in the following Eq.~\ref{eq:contrastive}.
\begin{equation}
\label{eq:contrastive}
\mathcal{L}_{MRL}(r_a, r_b)=
\begin{cases}
d_{ab}              & \text{if paired} \\
max(0,\ m - d_{ab}) & \text{if not paired}
\end{cases}
\end{equation}
where $d_{ab} = d(r_a, r_b)$.
Also, this solution uses a margin $m$ to stop moving away from the instances of different classes. 

\section{The proposed $\sigma^2$R Loss}
In this section we describe our proposed loss function called $\sigma^2$R loss, which aims to reduce the variance for each class, working in the feature space.
Starting from the well-known center loss function described in Eq.~\ref{equ:center_loss},
we introduce a multiplier $\beta\colon \mathbb{R}^d \to  \mathbb{R}$ based on a sigmoid function, to induce a weighted pumping state for each instance inside a batch of the training set: 
\begin{equation}\label{equ:beta}
    \beta(\sigma(n,x_{i}))=\frac{Z}{1+e^{-K \cdot (\sigma(n,x{_i}) - \sigma(n,C_{y_i}) )}}
\end{equation}
where 
$\sigma(n,x{_i})$ represents the standard deviation computed between $x_i$ and its $n$ neighbors, all belonging to the same class ${y_i}$,  $\sigma(n,C_{y_i})$ is the standard deviation considering only the class center $C_{y_i}$ and its $n$ nearest instances of the same class. 
Finally, $Z$ is a constant used to change the output range of the $\beta$ function from $[0,1]$ to $[0, Z]$ and $K$ is the logistic growth rate variable used to varying the slope of our function.
To avoid to find the optimal slope for the loss function, the $K$ parameter is automatically found by the learning process considering the following function (see Eq.~\ref{eq:growth_rate}) which reports the values in the range $[\epsilon,+\infty)$.
\begin{equation}\label{eq:growth_rate}
K(w_K)=\epsilon + \frac{Z}{1+e^{-w_K}}
\end{equation}
Experimentally we have found that the $K$ parameter is used a lot and the network changes it continuously and in Fig.~\ref{fig:k_growth} we report the trend of $w_K$ representing the growth rate values as the epochs vary.

\begin{figure*}
    \centering
    \subfloat[]{\includegraphics[height=3.5cm, width=.32\textwidth]{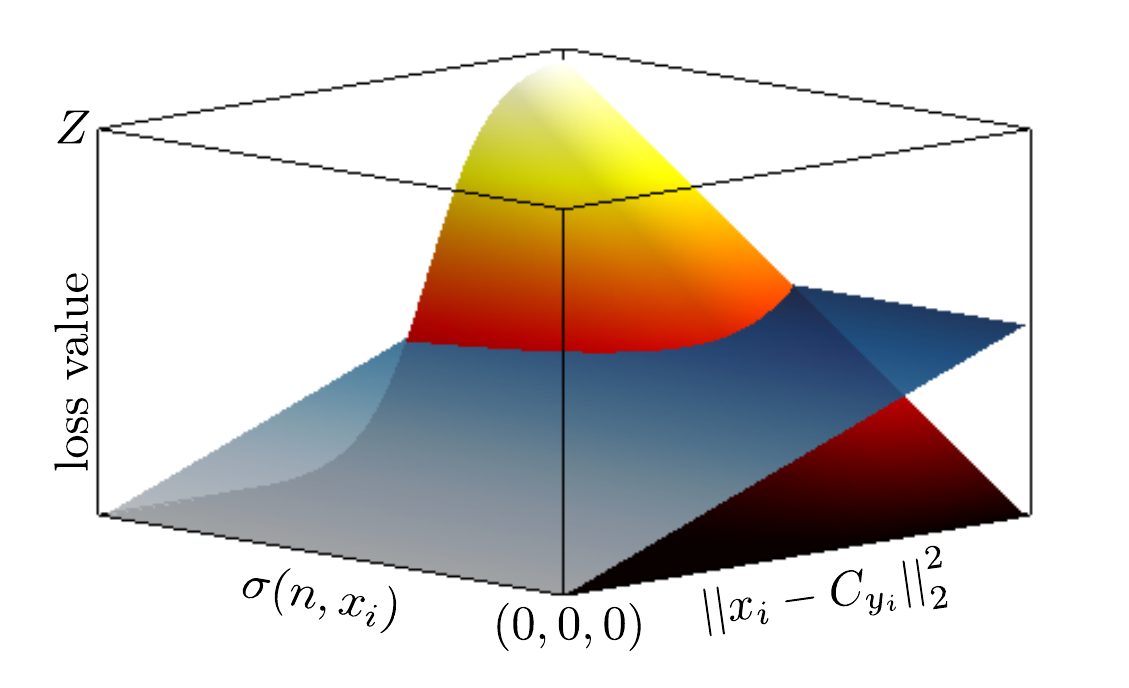}}
    \subfloat[]{\includegraphics[height=3.5cm, width=.32\textwidth]{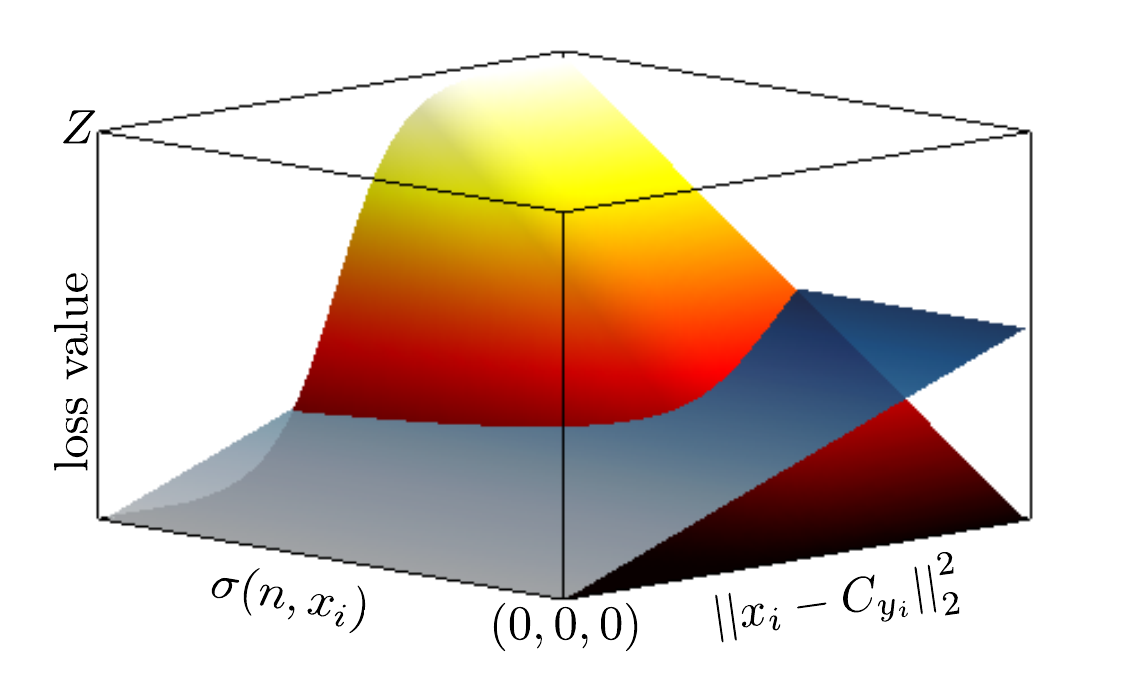}}
    \subfloat[]{\includegraphics[height=3.5cm, width=.32\textwidth]{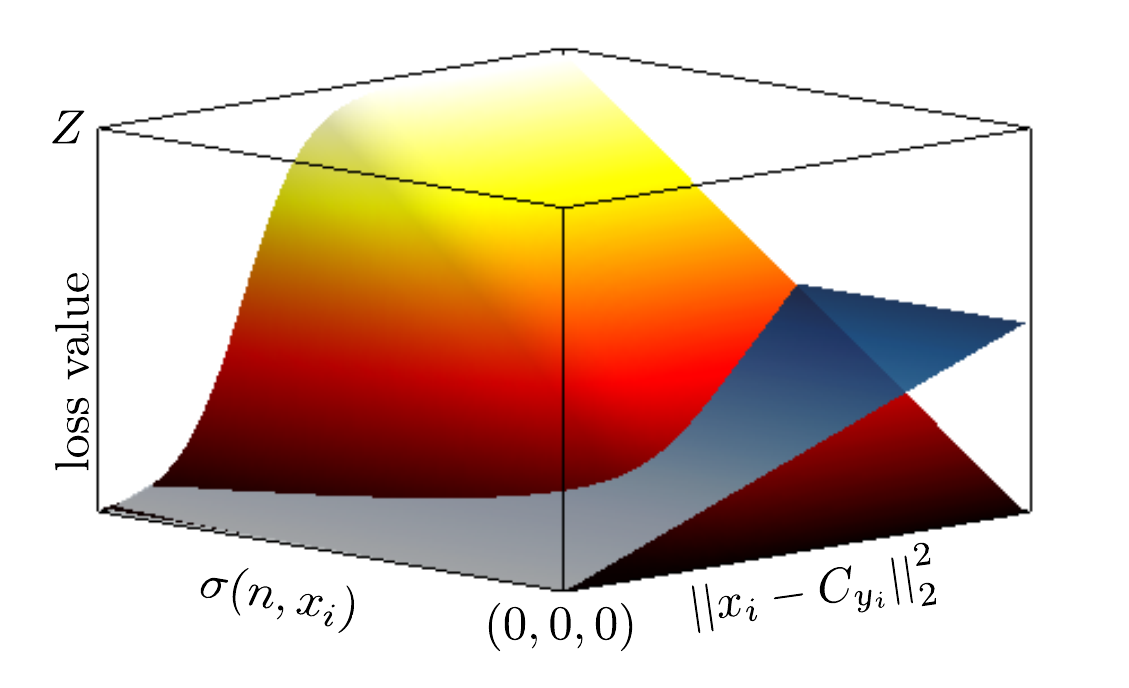}}
    \caption{In the three plots, we show the surface created by our loss function $\sigma^2$R and compare it with the central loss (blue/gray plane). The X-axis represents the Euclidean distance between the center of class $C_ {y_i}$ and the deep features of the same class. The Y-axis represents the standard deviation of the $n$ patterns closest to the sample $x_i$, while the Z-axis is the output range of the proposed loss function. In plots (a)-(c), we visualize the behavior of our function when the inflection point changes.} 
    \label{fig:surface}
\end{figure*}

Now we can define the proposed $L_{\sigma^2R}$ loss function as follow:
\begin{equation}\label{equ:pumping_center_loss}
    \mathcal{L}_{\sigma^2R}=\frac{1}{m} \sum\limits_{i=1}^m \beta(\sigma(n,x_i)) {|| x_i-C_{y_i}||}_2^2 
\end{equation}
where $m$ is the number of instances considered inside a training batch. 
$\beta$ is the particular sigmoid function having a growth rate variable $K$ and with an inflection point on ($\sigma(n, x {_i}) - \sigma(n, C_{ y_i}$), as defined in Eq.~\ref{equ:beta}.
A graphical representation of this loss function can be observed in Fig.\ref{fig:surface} where is possible to see the shape of the curve with three different distances between sample $x_i$ and class center.
An inflection point is defined as the point in which the function changes from being concave to convex or vice-versa. Consequently, we find the inflection point resolving the second derivative of the function considered. This point will be used as a reference to establish where the standard deviations of the nearest neighbours instances considered are located and then obtain the relative projections along the y-axis obtaining thus the increased error (see algorithm~\ref{alg:pseudocode}).
Leveraging on standard deviation, the idea is to pump all instances with their nearest neighbour less dense than the class center in a way to increase the error by a multiplier $\beta$ and in the same time to reduce the error of all instances with their nearest neighbours denser of the class center.
The loss function $L_{\sigma^2R}$ is minimized using a stochastic gradient descendent algorithm. 
In our approach, we do not consider only a single sigmoid with relative growth rate variable but we use a sigmoid for each class of the dataset used for the training step of the neural network. Growth rate variable $K$ and centroid variable $c_{y}$ are hyper-parameters in our proposal and will be handle by CNN used. We initialize them using a normal distribution (standard normal distribution) with mean $0$ and variance $1$.
To better understanding, we report the pseudocode in algorithm~\ref{alg:pseudocode}.

\begin{figure}
    \centering
        \includegraphics[scale=.38]{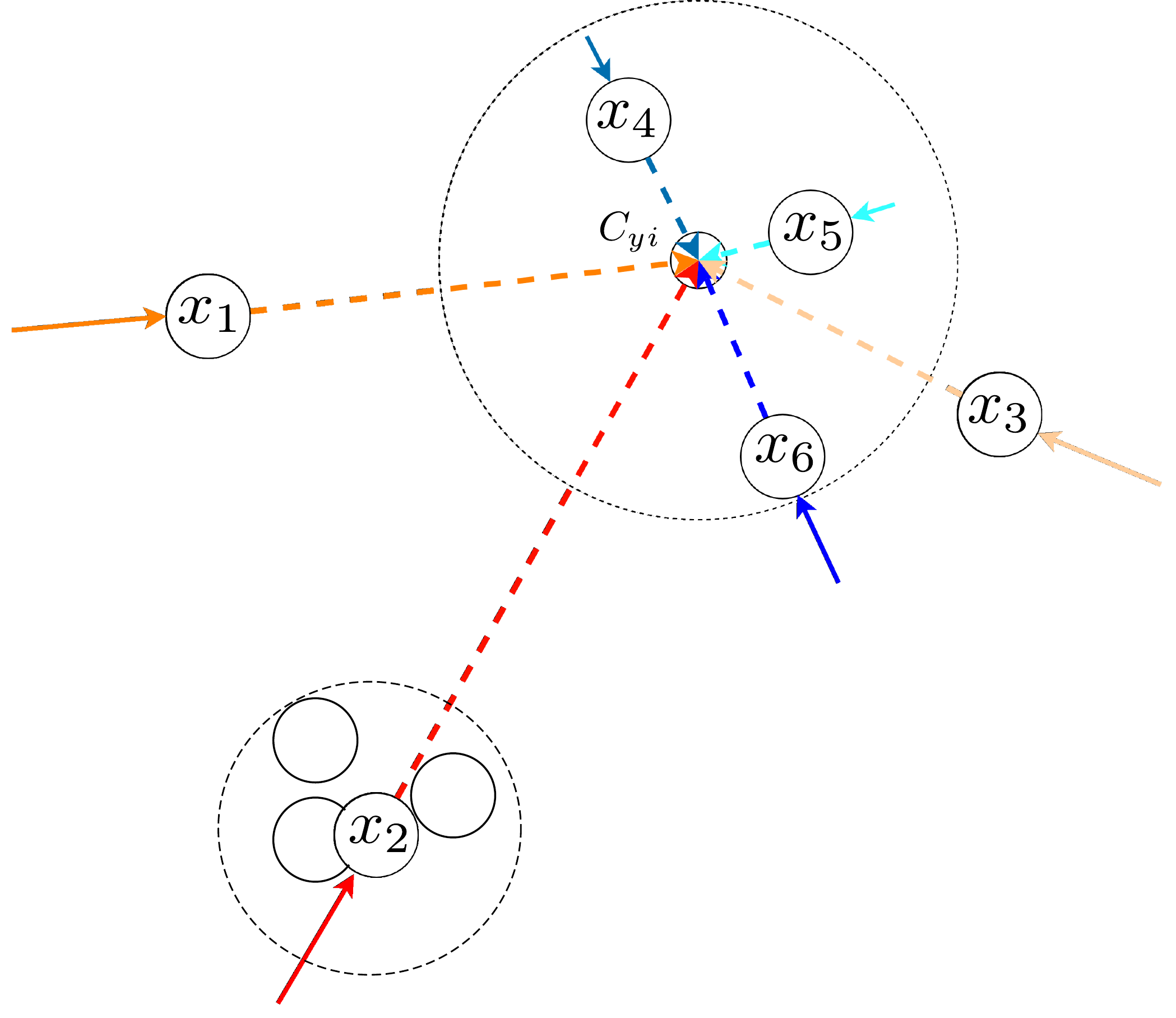}
    \caption{Graphical representation of some patterns $x_i$ in the deep features space when using the proposed $\sigma^2$R loss function.
All circles represent instances of the same class $y_i$ with their centroid $C_{y_i}$. The dotted circle is the boundary that contains the first $n=3$ closest samples to the class center. If you look at the patterns $x_1$, $x_2$ and $x_3$ which are the furthest away from the centroid $C_{y_i}$, their multiplier $\beta(\sigma(n,x_i))$ will be much larger than the multiplier $\beta(\sigma(n,x_i))$ of the patterns $x_4$, $x_5$ and $x_6$. 
Furthermore, comparing the furthest points $x_{1}$ and $x_{2}$ from the centroid, $\beta(\sigma(n,x_1))$ will be larger than $\beta(\sigma(n,x_2))$.
    }
    \label{FIG:main_idea}
\end{figure}



\begin{algorithm} \scriptsize
\caption{}
\begin{algorithmic}[1]
\Function{$\sigma^2$R Loss}{$x$, $y$}
\State {$m$} \Comment{$x$ contains $m$ samples and $y$ are the class labels}
\State {$\sigma(n,C_{y_i})$} \Comment{standard deviation computed on the $n$ points closest to $C_{y_i}$}

\For {$c_{j} \in {}$ \text{range($0$, $n\_classes$)}}
\State {$\sigma(n,x_i)$ \Comment{standard deviation computed on the $n$ points closest to $x_i \in c_j$}}
\State {$\beta(\sigma(n,x_i)) \gets \frac{Z}{1+e^{-K \cdot (\sigma(n,x{_i}) - \sigma(n,C_{y_i}) )}}
$}
\State {$ L_{c_j} \gets \sum\limits_{i=1,y_i = c_j}^{m} \beta(\sigma(n,x_i)) {|| x_i-C_{y_i}||}_2^2 $}
\State {$intra \gets intra + L_{c_j}$}
\EndFor \\
\State {$L_{\sigma^2R} \gets \frac{intra}{m}$}
\EndFunction
\end{algorithmic}
\label{alg:pseudocode}
\end{algorithm}






In Fig.~\ref{FIG:main_idea} we report a graphical representation of the effect obtained by applying the proposed $\sigma^2$R loss function.
As explained in the figure, the dashed circle is the boundary that contains the first $n$ closest samples to the class center $C_{y_i}$.
These samples are used to calculate the standard deviation so that the error generated by the more distant patterns can be inflated much more than the error generated by the closest patterns.
But there is a very important aspect to highlight by taking advantage of this example, comparing the furthest points $x_1$ and $x_2$  from the centroid, $\beta(\sigma(n,x_1))$ will be larger than $\beta(\sigma(n,x_2))$ because the standard deviation $\sigma(n,x_2)$ built on the neighborhood of $x_2$ is much smaller than the standard deviation $\sigma(n,x_1)$ built on the neighborhood of $x_1$. 
The proposed loss is low when the $n$ neighbours of an instance are denser than the neighbours of the centroid of a class and will be high when an instance is isolated and far from the class centroid.
Like the center loss function, the goal is to minimize the intra-class variance, but instead, to force the model to converge all the class samples in a single point, our loss function inflates/deflates the error to not have isolated instances or groups of instances with high standard deviation.

We emphasize that in Eq.~\ref{equ:beta} can have two different behaviours when the argument of exponential has a positive or negative value. This is due to the difference $\sigma(n, x {_i}) - \sigma(n, C_{ y_i})$ described before. 
As you can see on x-axis in Fig.~\ref{fig:surface}, when the distance is far from center of the class and $\sigma(n, x {_i}) - \sigma(n, C_{ y_i})>0$, our sigmoid assumes high values (red-orange color) thus we simulate the inflating error, by contrast our sigmoid will assume low values when the 
$\sigma(n, x {_i}) - \sigma(n, C_{ y_i})<0$ and in case the distance is very near to the center of the class, so as to simulate a frozen statement.
Finally, the total loss function we use in this paper must be considered jointly with the cross-entropy loss function as follow:
\begin{equation}\label{eq:full_loss}
    \mathcal{L}=\mathcal{L}_{xent} + \lambda \mathcal{L}_{\sigma^2R}
\end{equation}
where $\lambda$ is a scalar used to balance the two loss functions.

\subsection{A toy example}
\begin{figure}
    \centering
    {\includegraphics[width=0.7\columnwidth]{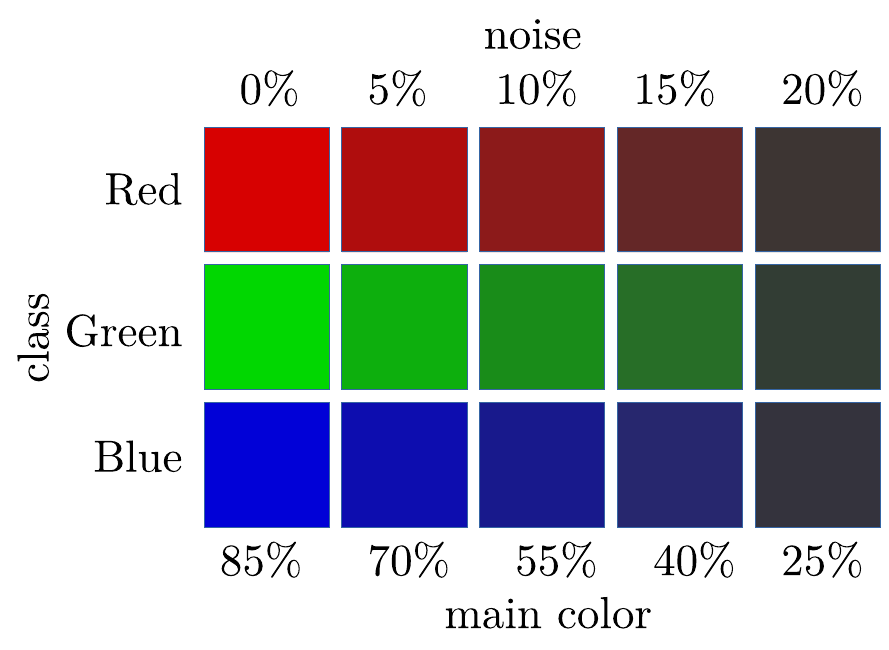}}
    \caption{
    Some samples of images extracted from the artificial image dataset \textit{Fuzzy-RGB} we created. It was obtained by combining different percentages of the three RGB channels. The three classes Red, Green, and Blue have a high value of "main colour" (between 100\% and 20\% of the maximum value) in the channel they represent and also contain a low percentage of "noise" (between 0\% and 20\%) of the two other channels.
    } 
    \label{fig:fuzzy_images_dataset}
\end{figure}

To better explain our loss, we introduce a toy dataset called \textit{Fuzzy-RGB} (see Fig.~\ref{fig:fuzzy_images_dataset}) having some characteristics that highlight the particular aspects of our loss function here proposed.
This dataset must have instances whose class attribution is uncertain, as showed in the rightmost column in Fig.~\ref{fig:fuzzy_images_dataset}.
With the same goal and in a similar way as done in~\cite{wen2016discriminative}, we reduce the second-last layer of a Resnet18 from 512 to 2 channels so that we can visualize the deep features in a two-dimensional space.
The comparative result between cross-entropy loss, cross-entropy plus center-loss, and $\sigma^2$R loss, applied to the toy dataset, is shown in Fig.~\ref{fig:dataset_toy}.
The results show us the maximization of the margin between the learned features of the three classes and the consequent better separability of the learned features, as well as the reduction of the variance for the learned features of each class (Fig.~\ref{fig:dataset_toy}c).
Analyzing the result of the model that uses cross-entropy plus the center loss function (Fig.~\ref{fig:dataset_toy}b) we can observe that, when many instances are close to the relative class center, the error will be very low, and consequently many instances will be left far from the center.
So, if the error is low, the learning process that uses cross-entropy plus the center loss function will stop the process of decreasing the distances between the features and their centroid, leaving everything in a frozen state.
For this reason, we proposed our loss function, which use a function to weigh the contribution of every single pattern to decrease the error and avoid the situation in which some points remain very far from their class centroid.

In Fig.~\ref{fig:fuzzy_images_dataset} we report 3 rows of images representative of the classes contained in the dataset used in this section to show the salient aspects of the proposed loss function. 
This Fuzzy-RGB dataset we propose is composed of RGB images of size $32\times32\times3 $ for a total of 3000 instances per class. 
Each image has a uniform RGB colour obtained by combining a high random value  of the "main colour" (between 100\% and 20\% of the maximum value) in the channel they represent and also a low percentage of random "noise" (between 0\% and 20\%) of the two other channels.

\subsection{Intra-class measure}
Within each class, the variance is a measure of variability widely used in deep learning to control the average distance between the samples of a class in the deep features space. 
We use the following notation to calculate the intra-class measure:
\begin{equation}\label{eq:intra}
\displaystyle I_{c_j}=\sqrt{\frac{\sum_{x_{i}\in c_{j}}(x_{i}- \mu)^2}{(n - 1)}}
\end{equation}
where $c_j$ is the class of the $x_i$-instance and $\mu$ is the mean value of the class $c_j$ and $n=|c_j|$.
Intra-class is widely used to measures how far a set of samples are spread out from their average value. 

\section{Datasets}
To demonstrate the effectiveness of our proposed method we use some different classification datasets.

\textbf{Cifar10 and Cifar100}~\cite{krizhevsky2009learning} are datasets commonly used as benchmarks in the literature. 
They both have 60,000 training samples and 10,000 test samples, and each input is a $32 \times  32$ RGB image. 
Cifar10 and Cifar100 have 10 and 100 classes respectively.

\textbf{FashionMnist}~\cite{xiao2017fashion} is a dataset consisting of 60,000 training samples and 10,000 test samples. 
Each image is $28  \times  28$ in grayscale and is associated with a label among the 10 different classes.
It is used as a reference dataset by many algorithms to compare with other papers in the literature.

\begin{table}[width=.9\linewidth,cols=4,pos=h]
\caption{Intra-class variance $I_{c_j}$ comparison between center loss and $\sigma^2$R loss on FMNIST dataset. We set $\lambda$=0.01, learning rate of the model to 0.4 and learning rate for center loss and our proposal equals to 0.1, batch size 256.}
\label{tab:intravarianceFMNIST}
\text{Train}
\begin{tabular}{l|c|c|c }
\toprule
Class  & Center Loss & $\sigma^2$R Loss & $\delta$ \% \\ \midrule
Class 0        &    0.8378    &    0.1904    &    339.87 \\
Class 1        &    1.0190    &    0.2155    &    372.65 \\
Class 2        &    1.0543    &    0.2628    &    301.05 \\
Class 3        &    1.0234    &    0.2289    &    347.01 \\
Class 4        &    1.0023    &    0.2406    &    316.53 \\
Class 5        &    0.8798    &    0.2026    &    334.11 \\
Class 6        &    1.1142    &    0.2785    &    300.03 \\
Class 7        &    0.9149    &    0.1911    &    378.74 \\
Class 8        &    1.0085    &    0.2462    &    309.48 \\
Class 9        &    0.7721    &    0.1579    &    388.83 \\
\end{tabular}
\\
\text{Test }
\begin{tabular}{l|c|c|c }
\toprule
Class  & Center Loss & $\sigma^2$R Loss & $\delta \%$ \\ \midrule
Class 0        &    0.7372    &    0.1559    &    372.87 \\
Class 1        &    0.9101    &    0.1759    &    417.45 \\
Class 2        &    1.0292    &    0.2661    &    286.70 \\
Class 3        &    1.0061    &    0.2159    &    365.89 \\
Class 4        &    0.9773    &    0.2354    &    315.14 \\
Class 5        &    0.7601    &    0.1638    &    363.80 \\
Class 6        &    1.1069    &    0.2743    &    303.46 \\
Class 7        &    0.7824    &    0.1556    &    402.56 \\
Class 8        &    0.9867    &    0.2361    &    317.90 \\
Class 9        &    0.6187    &    0.1335    &    363.42 \\
\bottomrule
\end{tabular}
\end{table}

\section{Experiments}
Now we present our experimental results.
Before starting with the description of the experiments, we group the settings of the various experiments in the following section.
Below, the experiments have been grouped into two main groups, in the first group we compare the proposed loss function with the center loss and the cross-entropy, in terms of classification accuracy.
In the second group of experiments, we compare our loss with the center loss functions and the cross-entropy, analyzing the intra-class variance in detail to highlight the relevant differences.

\begin{figure*}
    \centering
    \subfloat[]{\includegraphics[height=4cm]{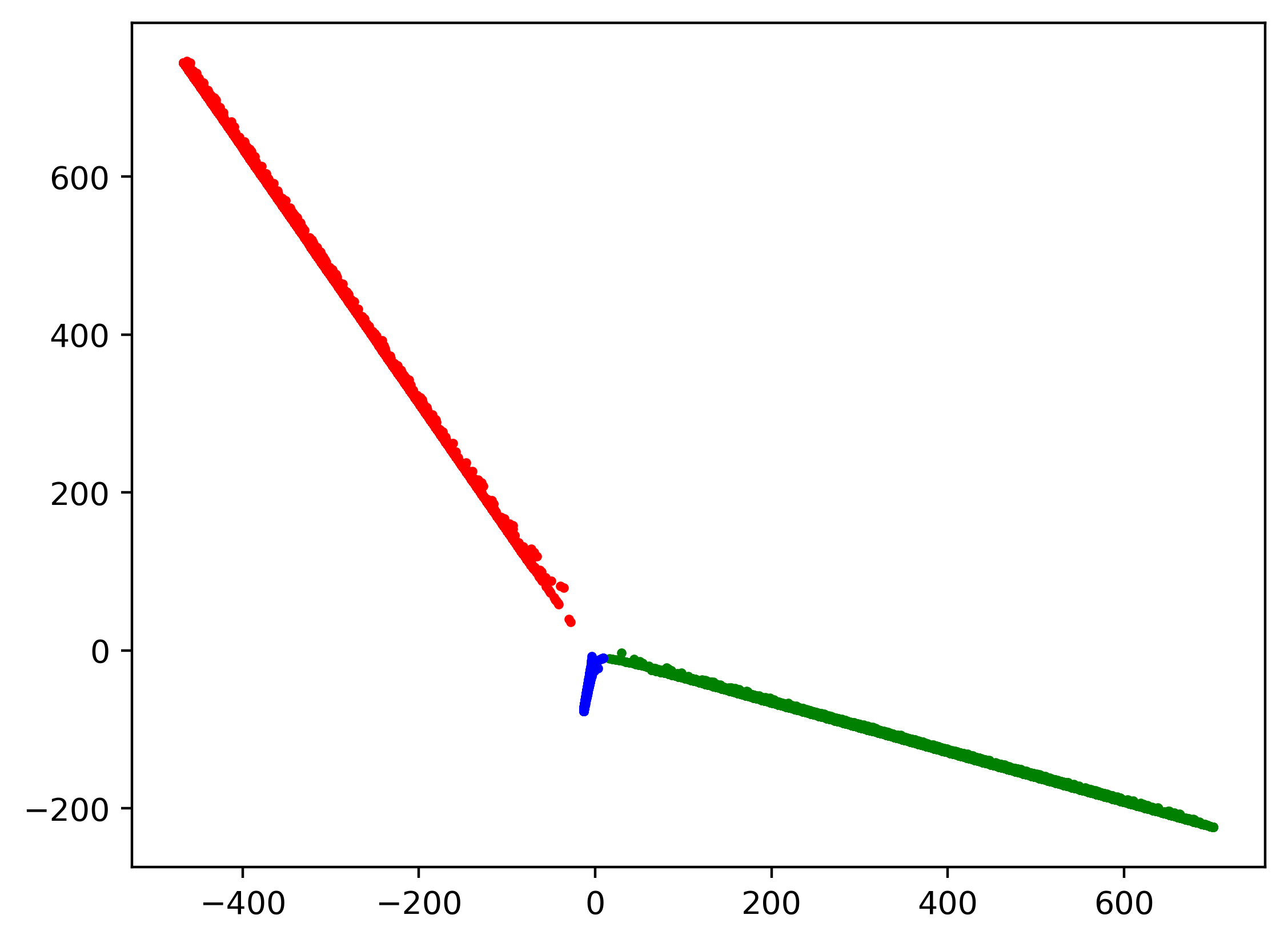}}
    \subfloat[]{\includegraphics[height=4cm]{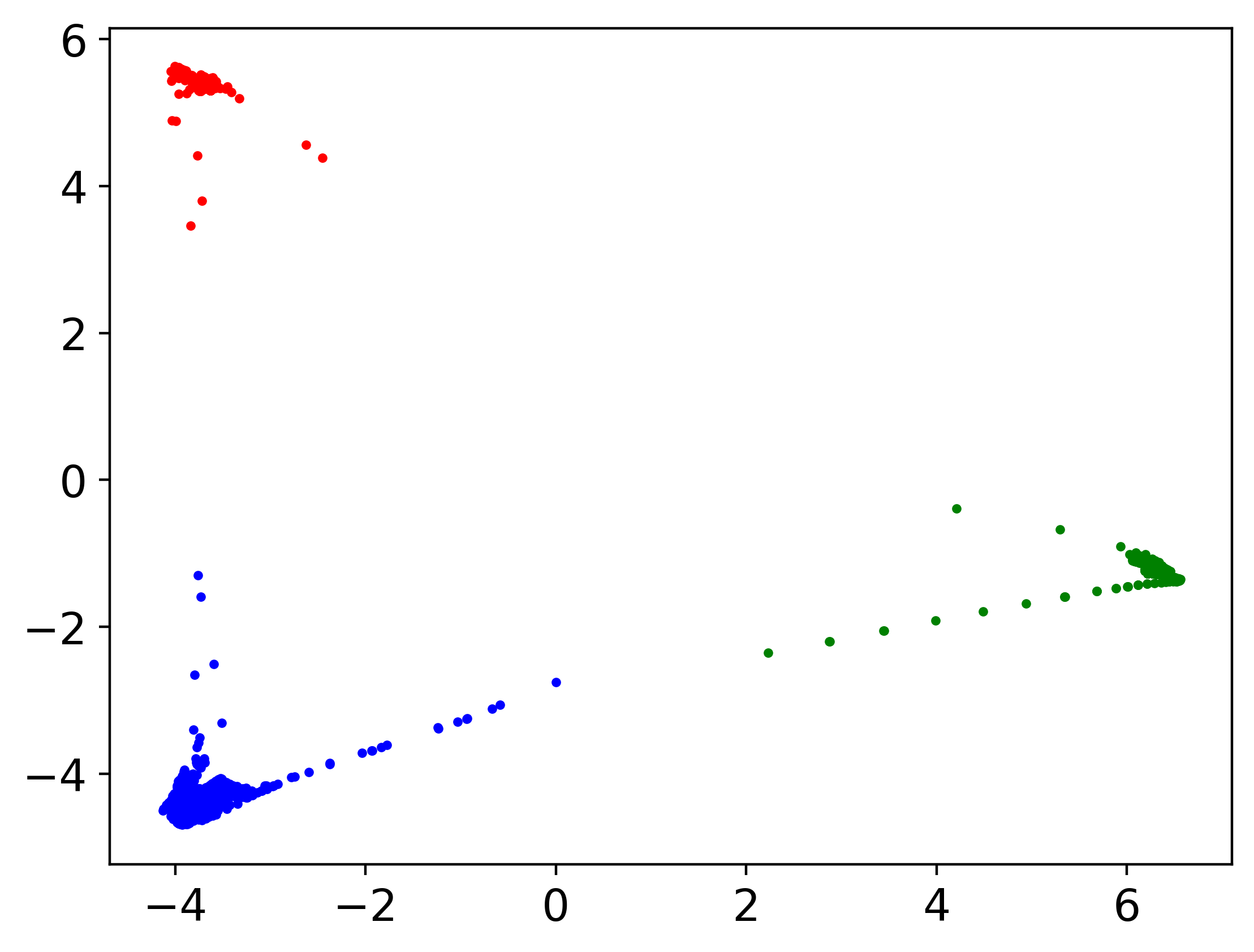}}
    \subfloat[]{\includegraphics[height=4cm]{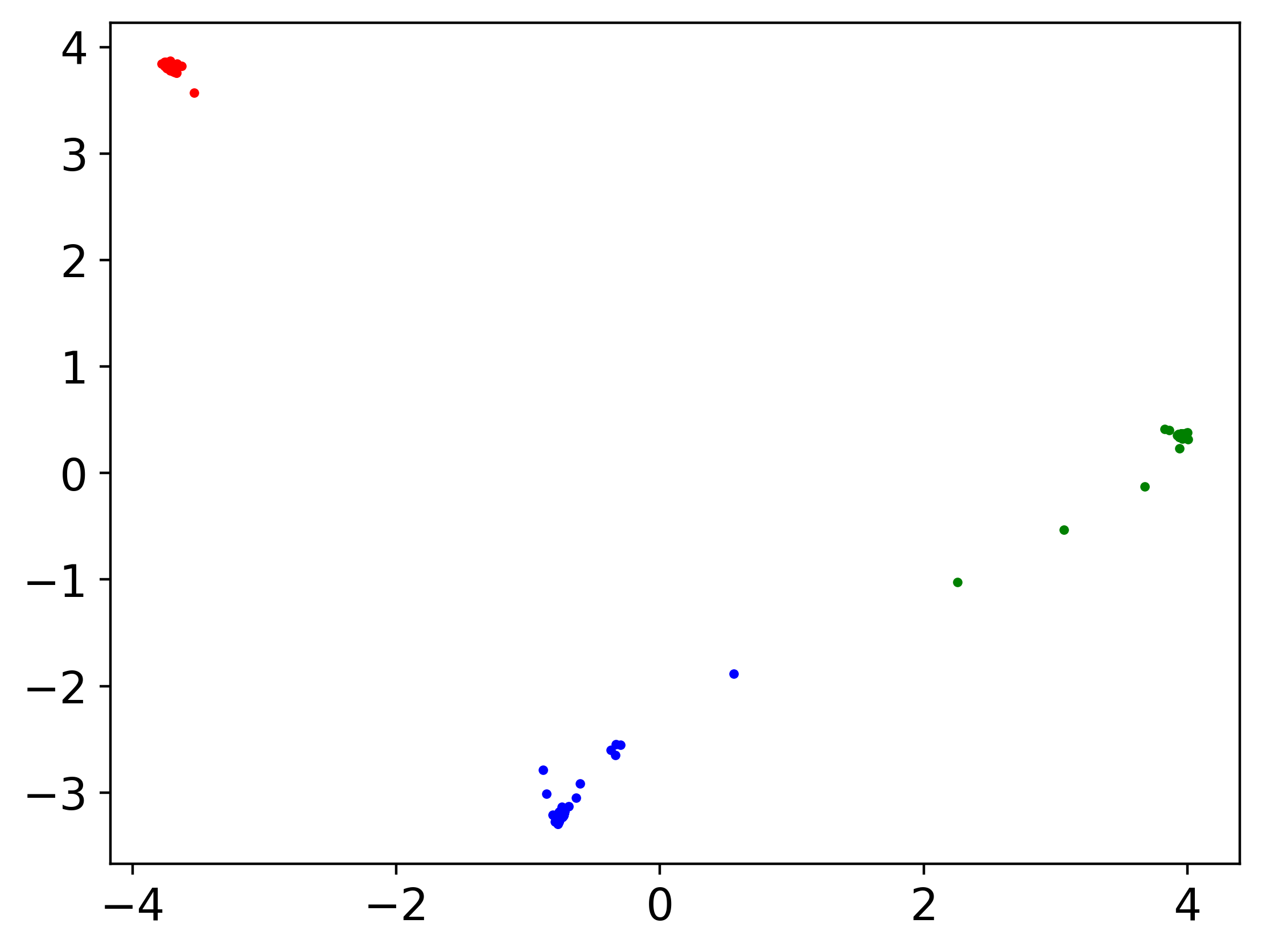}}
    \caption{Quality comparison on the training set, between the $\sigma^2$R loss function and two other loss functions on the Fuzzy-RGB dataset we created.
In (a) the 2D representation of the deep features found by a LeNet with PReLU activation function that uses the cross-entropy function, in (b) the same network trained with the center loss function and in (c) the deep features arranged around the class centroids thanks to the work of our loss function. To view the deep features, CNN's penultimate layer uses only two neurons. Figures (b) and (c) show similar behaviour but in (c) the cluster around the class centroid is much more compact.}
\label{fig:dataset_toy}
\end{figure*}

\subsection{Network settings}
We have implemented $\sigma^2$R loss in Pytorch~\cite{paszke2019pytorch} using Python3 as the programming language. 
The neural models used as baselines are Resnet18 and LeNet.
All models were trained using an Nvidia Titan X GPU.
As for the learning rate, we used an adaptive learning decay, and in particular, we used a cosine-like function to reduce the learning rate at each epoch.
We use two Adam~\cite{kingma2014adam} optimizers that allow us to update network weights and loss function parameters iterative based on training data.
In particular, we use an Adam optimizer to find the best weights of the neural model and a second Adam optimizer for the parameters of our loss function and also for the center loss.
To highlight the characteristics of our loss function, we have decided to increase the amount of data. For this reason, we use algorithms capable of generating synthetic data similar to the real ones, through procedures that require to be able to reproduce both realistic noise and imitate all possible variations of the samples in the real world. 
We used various strategies such as random cropping, random rotation, and vertical/horizontal flipping to expand an existing dataset to form a model over multiple examples.
After that, we re-scale each image to 32$\times$32$\times$3 pixels (except for FashionMnist where we have only 28$\times$28$\times$1 pixels).
We emphasize that we used the original Resnet18 and LeNet, without specific modifications that allowed us to obtain the best performance on the specific datasets used.
In each run, we have a random and therefore different initialization of the network weights, but we decided to set the same order in which the data are sampled by the network to have better comparisons in our experiments.
In addition, we apply a balanced data sampler to extract each batch due to the nature of our proposal and in order to have a batch of samples balanced across all classes.

\subsection{Experiment 1}
In the first experiment, we compare three different loss functions using standard datasets used in other papers so that we can compare our proposal with the solutions present in the literature. 
In particular, we compared the cross-entropy loss function, the center loss and our loss function using a Resnet18~\cite{He_2016} as a neural model applied to the Cifar10 and Cifar100 datasets. Furthermore, we also used the LeNet CNN on the FMNIST dataset following the setting described in~\cite{frosst2019analyzing}.
During the training of neural models, for each epoch, we calculate the intra-class variance of the training class and the test class, computed as standard deviation according to the formula $I_ {c_j}$ reported in Eq.~\ref{eq:intra}.
We use this metric to check for variance within the class (intra-class) and this is a very important concept for checking the separation between classes, attracting elements of the same class.
Furthermore, we also check the intra-class variance in the test phase to observe the correct generalization ability by the model in attracting instances within the same class for never seen instances.

\begin{figure*}
    \centering
    \subfloat[]{\includegraphics[height=5cm, width=5.3cm]{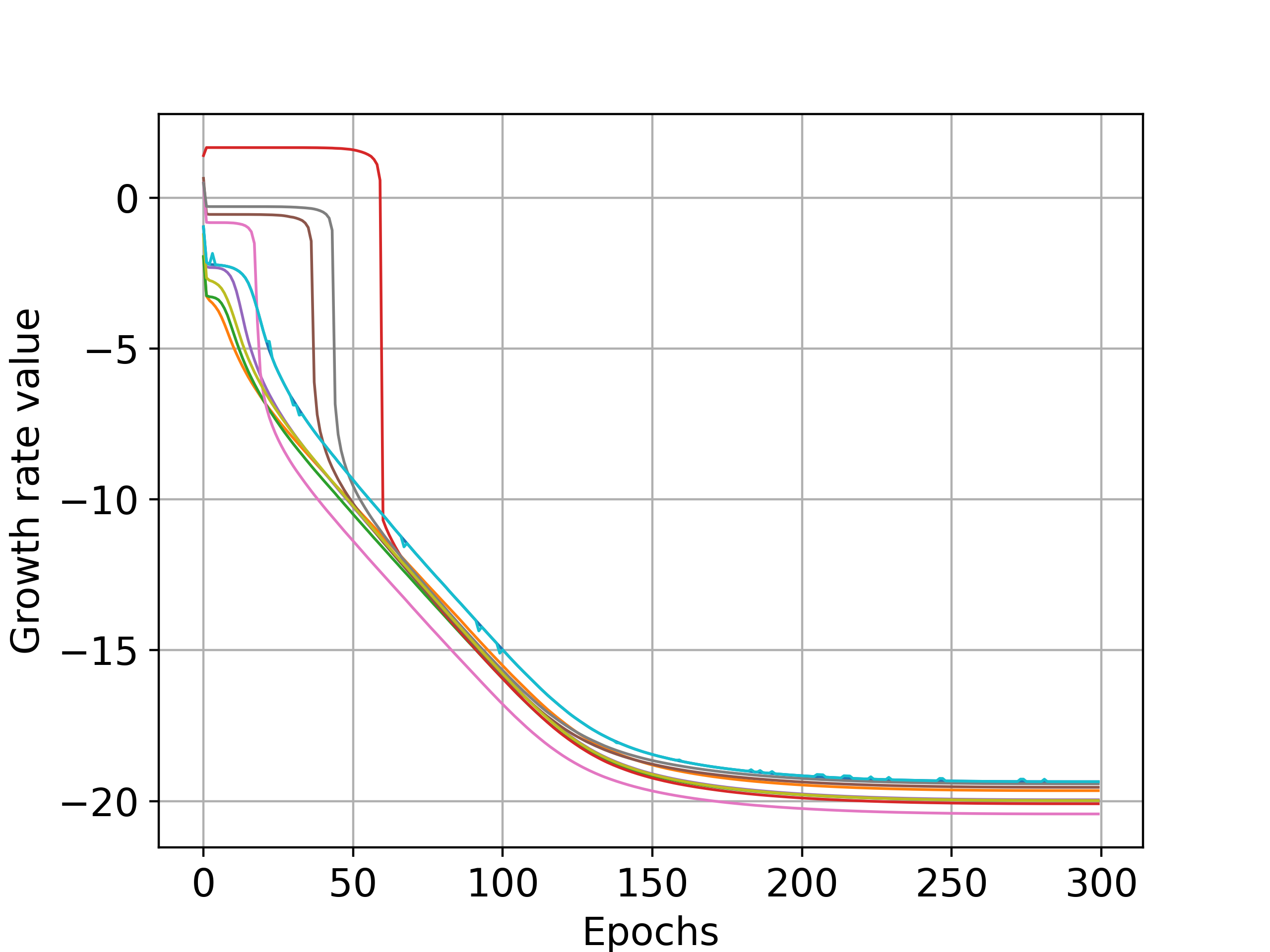}}
    \subfloat[]{\includegraphics[height=5cm, width=5.3cm]{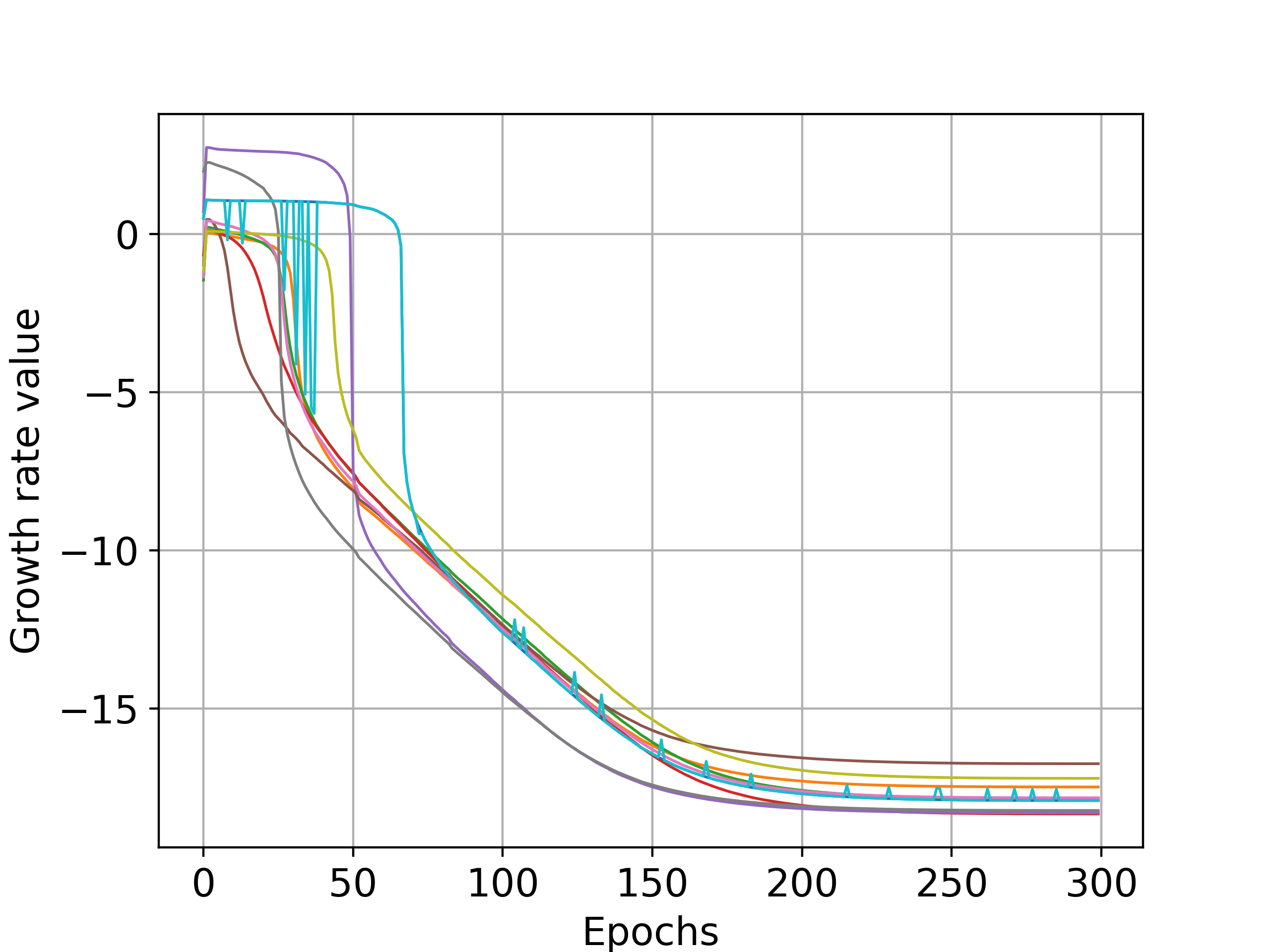}}
    \subfloat[]{\includegraphics[height=4.6cm, width=5.3cm]{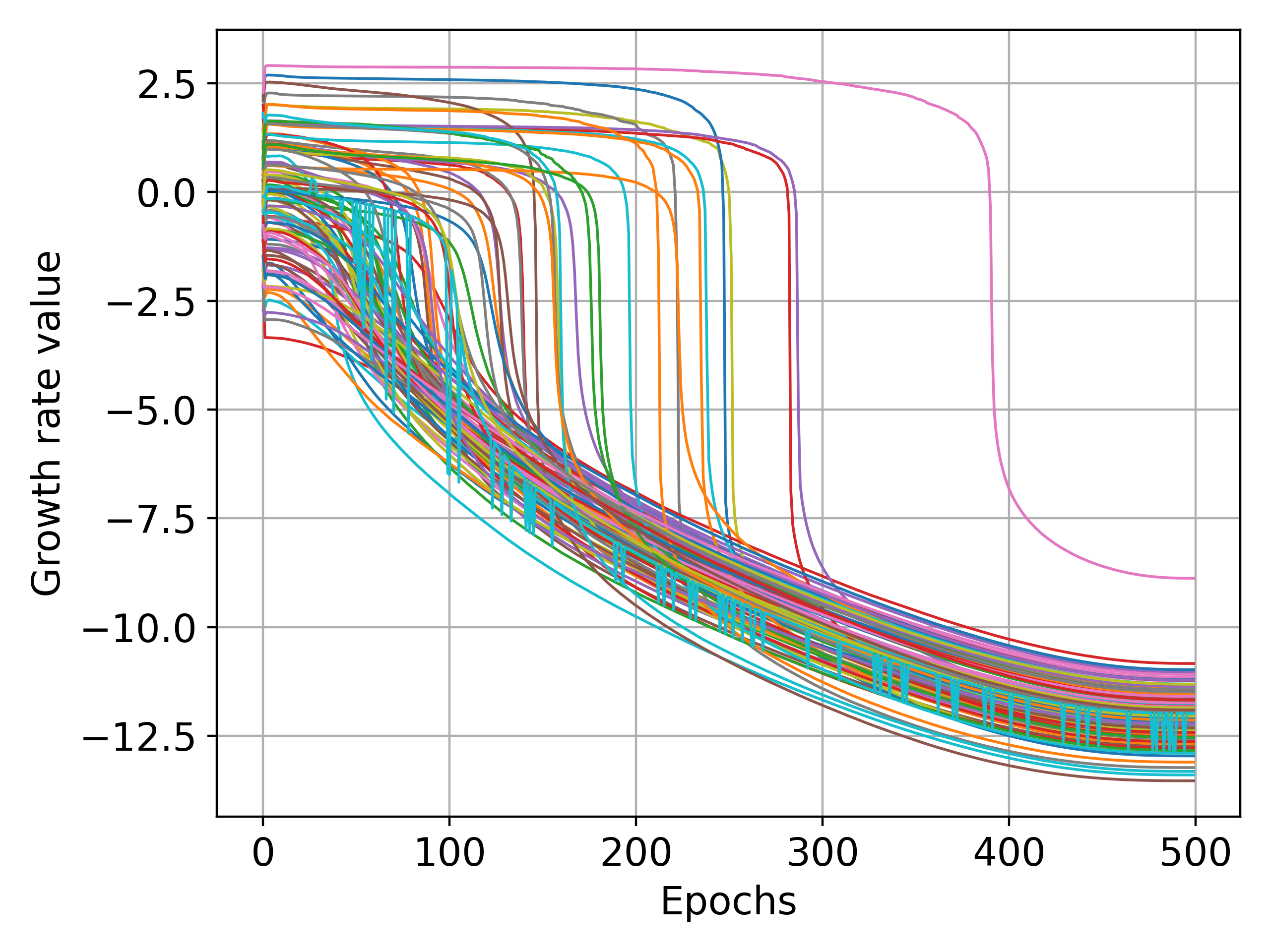}}
    \caption{The three figures show the change in the growth rate value $w_K$ described in Eq.~\ref{eq:growth_rate}, as it is learned by the neural model during the training phase. In (a) the behavior obtained on the FMNIST dataset, in (b) on Cifar10 and in (c) on the Cifar100. The number of plots in each figure coincides with the number of classes of the dataset used, therefore each plot represents the relative growth variable $w_K$ associated with a specific class.
    } 
    \label{fig:k_growth}
\end{figure*}

In Tabs.~\ref{tab:hinton_comparison} and~\ref{tab:cifar100} we show the effectiveness of our loss function in terms of accuracy, comparing ourselves with the same models that use the cross-entropy loss function, the center loss plus the cross-entropy loss function or the SoftN loss function proposed in the paper~\cite{frosst2019analyzing}.
In these tables, we report the average and maximum accuracy of 10 executions for all datasets.
The ResNet18 model was trained for 300 epochs using the Cifar10 dataset and was trained for 500 epochs using the Cifar100 dataset.
The $\lambda$ parameter has been set equal to 0.01 for both loss functions. 
Due to the characteristics of our methodology which exploits the proposed loss function, the batch sizes are set respectively to 256 and 1000 for Cifar10 and Cifar100. Furthermore, again for the same reason, we use balanced batches when loading the dataset, therefore each batch will have about 25 instances per class for the Cifar10 and FMNIST datasets which have 10 classes and the batch size has been set to 256, while there are 10 instances per class for the Cifar100 dataset. The main reason why we need this last constraint is in the computation of the standard deviation introduced In Eq.~\ref{equ:beta}, for which we must avoid having a situation in which for example a batch has 0 instances for a specific class or with fewer instances than the chosen neighbourhood constraint $n$.
The neighbourhood constraint was set with a parameter $n=7$ and the output range $Z$ of the $\beta$ function was set to 40 (chosen arbitrarily).
In Tabs.~\ref{tab:hinton_comparison} and~\ref{tab:cifar100} we report the results in terms of accuracy. In the same tables, it is possible to notice that on Cifar10 and Cifar100 we exceed the performances of the configurations that do not use our loss function (the baselines reported in tables) and we also report the percentage gain using our loss function compared to cross-entropy, center loss + cross-entropy and the loss proposed by~\cite{frosst2019analyzing}.
For the experiment carried out on the FMNIST dataset, we used a LeNet with the same parameters as the experiment conducted on the Cifar10 dataset except for the learning rate associated to the optimizer which was set to 0.001.
All other parameters are the same as those used in the paper~\cite{frosst2019analyzing}.
In this experiment, our loss function shows better performance than the center loss function, the cross-entropy loss function and the loss function proposed in~\cite {frosst2019analyzing}.

From this first group of experiments, we can conclude that our loss function certainly contributes to improving the classification accuracy if compared with the same model that uses only cross-entropy as a loss function and both comparing it with the same model that uses the center loss function.

\begin{table}[width=.9\linewidth,cols=4,pos=h]
\caption{Intra-class variance comparison on Cifar10 dataset between center loss and $\sigma^2$R loss. 
To compute these values we set $\lambda$=0.01, learning rate of the neural model equals to 0.4 and learning rate for center loss and our proposal equals to 0.1, batch size equals to 256.}
\label{tab:intravariancecifar10}
\begin{tabular*}{\tblwidth}{@{} L|C|C|C@{} }
\toprule
Cifar10  & Center Loss & $\sigma^2$R Loss & $\delta$ \% \\
\midrule
Class    0    &    0.4127    &    0.1016    &    306.12    \\
Class    1    &    0.2450    &    0.0698    &    250.86    \\
Class    2    &    0.4062    &    0.1093    &    271.74    \\
Class    3    &    0.2800    &    0.1452    &    92.88  \\
Class    4    &    0.3656    &    0.0991    &    268.79    \\
Class    5    &    0.3575    &    0.0967    &    269.76    \\
Class    6    &    0.3067    &    0.0819    &    274.54    \\
Class    7    &    0.2826    &    0.0879    &    221.56    \\
Class    8    &    0.3197    &    0.0712    &    349.25    \\
Class    9    &    0.3074    &    0.0798    &    285.15    \\
\bottomrule
\end{tabular*}
\end{table}

\begin{table}
\caption{Accuracy benchmark comparison on Cifar10 using our loss and soft nearest neighbor (SoftN) loss~\cite{frosst2019analyzing}. 
For all models we use a ResNet18 with learning rate of the neural model equals to 0.4 and learning rate for center loss and our proposal equal to 0.1, Adam optimizer and batch-size equals to 256. 
The table shows the average accuracy over 10 executions (first column), the maximum accuracy over 10 executions (second column) and the gain $\delta$ in percentage, comparing the $\sigma^2$R loss function and the cross-entropy baseline (the same model using the cross-entropy loss function) and soft neighbour loss with the same baseline in the third column. Furthermore, we compare the $\sigma^2$R loss and center loss in the last two rows (here the baseline is the cross-entropy plus the center loss function).
In all experiments, we obtain a greater gain than cross-entropy, center loss, and in comparison with~\cite{frosst2019analyzing}.
}\label{tab:hinton_comparison}
\begin{tabular*}{\tblwidth}{C|C|C|C}
\toprule
& $\sigma^2$R Loss  & baseline & $\delta\%$ \\
\midrule
avg    &    \textbf{91.503}    &    89.71    &    1.959 \\
max    &    \textbf{92.01}    &    90.78    &    1.336 \\
\hline
& SoftN~\cite{frosst2019analyzing} & baseline  & $\delta\%$ \\
\hline
avg    &    89.9    &    89.71    &    0.211 \\
max    &    91.22    &    90.78    &    0.482 \\
\hline
& $\sigma^2$R Loss  & base+CenterLoss & $\delta\%$ \\
\hline
avg    &    \textbf{91.503}    &    91.34    &    0.178 \\
max    &    \textbf{92.01}    &    91.72    &    0.315 \\
\bottomrule
\end{tabular*}
\end{table}

\subsection{Experiment 2}
To better visualize and highlight the characteristics of our $\sigma^2$R loss function and also be able to quantify the improvements obtained in terms of numerical values, in this experiment we show the intra-class variance obtained with the training set on our Fuzzy-RGB dataset.
In this experiment, we used a ResNet18 in which we changed the size of the second-last layer from 512 channels to 2 channels, so that the deep features of the Fuzzy-RGB problem could be viewed in a 2D space.
In Fig.~\ref{fig:dataset_toy} three plots are shown from left to right show the arrangement of the deep features for the same ResNet18 trained using the cross-entropy loss function, the center loss function and our $\sigma ^2$R loss function.
As it is easy to understand from the figure, using our loss function we obtain an incredible improvement in terms of reduction of intra-class variance in the training set.
In fact, by observing the instances of the three classes of the problem, it is clearly seen that these are positioned very close to their class centroid, although some of these instances are very noisy and therefore the class to which they belong is ambiguous.

\begin{table}
\caption{Accuracy benchmark comparison on Cifar100 using $\sigma^2$R Loss, cross-entropy (baseline) and cross-entropy+center loss. As in Cifar10, we report a better accuracy than baseline and baseline+CenterLoss.
}\label{tab:cifar100}
\begin{tabular*}{\tblwidth}{C|C|C|C}
\toprule
& $\sigma^2$R Loss  & baseline & $\delta\%$ \\
\midrule
avg    &    \textbf{60.281}    &    56.44    &    6.371    \\
max    &    \textbf{62.65}    &    58.18    &    7.134    \\ \hline 
&    $\sigma^2$R Loss        &    base+CenterLoss    &    $\delta\%$    \\
\hline                            
avg    &    \textbf{60.281}    &    58.90    &    2.284    \\
max    &    \textbf{62.65}    &    61.57    &    1.723    \\
\bottomrule
\end{tabular*}
\end{table}

To numerically quantify the difference in terms of intra-class variance, we have performed two new experiments on the Cifar10 and FMNIST datasets.
The 4-\textit{th} column in Tab.~\ref{tab:intravariancecifar10} shows the percentage change in our loss function compared to the center loss function and shows an increase in the compactness of the deep features of each class, with percentage improvements that vary from 92\% to 349\% on Cifar10.
A similar experiment is shown in Tab.~\ref{tab:intravarianceFMNIST} and shows the intra-class variance measure in training and testing for the FMNIST dataset.

Finally, applying a ResNet18 on the datasets Cifar10, Cifar100, and a LeNet on FMNIST, we analyzed the variability during the training phase of the parameter $w_K$ described in Eq.~\ref{eq:growth_rate}.
The behaviours obtained on the three datasets are shown in Fig.~\ref{fig:k_growth}. 
In each plot of this figure, the curves represent the relative growth variable $w_K$ assigned to a specific class of the dataset used. 
In the y-axis the values assigned to $w_K$ at each epoch from the learning process.
The learned $w_K$ values are then normalized using a sigmoidal function to bring them back into a range $[0,1]$ and thus avoid negative values, and finally, we move from the range $[0,1]$ to the range [$\epsilon$, 40] multiplying by 40 (arbitrarily chosen value) and adding a $\epsilon$ to avoid zero.
As can be seen from the Fig.~\ref{fig:k_growth}, after the initialization of the $w_K$ parameter with values close to zero, the learning process takes this parameter to assume smaller values which depend on the dataset used. 
A small value of the $w_K$ parameter corresponds to a sigmoid defined in Eq.~\ref{eq:growth_rate} with a very low value. Note that this final effect comes only at the end of the learning process, which shows that having a sigmoid for this parameter has its usefulness, especially during the initial phase of the learning process.


\section{Conclusions}
In this paper, we have introduced a new loss function to reduce furthermore the intra-class variance and overcome the center loss function performances. 

The final error of a center loss is strongly influenced by the majority of instances that are close to the centroid but it is not affected by isolated points that are far from the class center and it can bring the model to a frozen state in terms of intra-class variance efficiency.
To tackle this problem, we use a weighted approach using sigmoid functions introduced as learning parameters of a neural network to pump or freeze the error based on squared Euclidean distance for each instance of the training set. 
The $\sigma^2$R loss has a clear intuition and geometric interpretation as we showed in the paper. 
Extensive experiments on several benchmark datasets demonstrate the effectiveness and usability of the proposed loss.
Future works plan to make scalable our loss function in a way to conduct experiments on a huge dataset having a large number of classes as Imagenet and to investigate different functions that will change the main behaviour to inflate/deflate the error.
\bibliographystyle{main}
\bibliography{main}
\end{document}